\documentclass[a4paper,10pt,twocolumn]{article}
\usepackage{balance}
\usepackage{amsmath}
\usepackage{mathrsfs}
\usepackage{amsfonts}
\usepackage{amssymb}
\usepackage{amsthm}
\usepackage{sectsty}
\sectionfont{\large}
\subsectionfont{\normalsize}
\usepackage[top=3cm, bottom=2.5cm, left=2cm, right=2cm]{geometry}
\DeclareMathOperator*{\nut}{arg}
\newtheorem{theorem}{Theorem}

\newtheorem{proposition}{Proposition}

\newtheorem{assumption}{Assumption}

\title{\textbf{\Large Eigenvectors for clustering: \\Unipartite, bipartite, and directed graph cases}}
\author{\large Andri Mirzal and Masashi Furukawa\\ \normalsize Graduate School of Information Science and Technology,\\ \normalsize Hokkaido University, Kita 14 Nishi 9, Kita-Ku,\\
\normalsize Sapporo 060-0814, Japan}
\date{}
\begin{document}

%Start the page number with 1
\setcounter{page}{1}
\maketitle

\noindent\textbf{Abstract}: This paper presents a concise tutorial on spectral clustering for broad spectrum graphs which include unipartite (undirected) graph, bipartite graph, and directed graph. We show how to transform bipartite graph and directed graph into corresponding unipartite graph, therefore allowing a unified treatment to all cases. In bipartite graph, we show that the relaxed solution to the $K$-way co-clustering can be found by computing the left and right eigenvectors of the data matrix. This gives a theoretical basis for $K$-way spectral co-clustering algorithms proposed in the literatures. We also show that solving row and column co-clustering is equivalent to solving row and column clustering separately, thus giving a theoretical support for the claim: ``column clustering implies row clustering and vice versa''. And in the last part, we generalize the Ky Fan theorem---which is the central theorem for explaining spectral clustering---to rectangular complex matrix motivated by the results from bipartite graph analysis.\\
\textbf{Keywords}: eigenvectors, graph clustering, Ky Fan theorem, spectral methods.

\section{Introduction} \label{introduction}
Many papers have been written to reveal the secret and power of spectral clustering; the using of eigenvectors of an af\mbox{}finity matrix induced from a graph to find natural grouping of the vertices. Some noteworthy works are \cite{Zha, Ding, Shi, Ng, Yu}, and a comprehensive tutorial can be found in \cite{Luxburg}. Despite being intensively studied, it is quite hard to find an intuitive and concise explanation on how and why the spectral clustering works. So, the logic behind the spectral clustering will be explained first.

Most works deal with bipartite data clustering since many real datasets such as a collection of documents, movie ratings, and experimental samples are bipartite. The usual approach for this case is to transform the feature-by-item rectangular matrix induced from a bipartite dataset into a corresponding symmetric matrix by using a kernel function. Then, a similar treatment as in unipartite graph can be employed to this symmetric matrix to find the clusters.

However, simultaneous row and column clustering (co-clustering) works in the original data matrix, hence, the above approach will not work. In subsection \ref{bipartite} we show that the co-clustering problem can be restated into the clustering of bipartite graph with two type of vertices---item vertices and feature vertices---where the induced af\mbox{}finity matrix is symmetric. Thus, various clustering algorithms built for unipartite graph can be employed directly.

In directed graph, usually edge directions are ignored to get an equivalent unipartite graph representation. However as noted in \cite{Leicht}, ignoring the edge directions can lead to a poor result, and a significant improvement can be achieved by counting for the edge directions into the model. As rows and columns of the induced af\mbox{}finity matrix of a directed graph correspond to the same set of vertices with the same order, as long as the clustering problem is concerned, a symmetric matrix can be formed by simply adding the matrix to its transpose. Therefore, allowing similar treatment as in unipartite graph. We will discuss this more details in subsection \ref{directed}.

A note on notation. $\mathbb{C}^{N\times K}$ denotes an $N\times K$ complex matrix, $\mathbb{R}^{N\times K}$ denotes an $N\times K$ real matrix, $\mathbb{R}_{+}^{N\times K}$ denotes an $N\times K$ nonnegative real matrix, $\mathbb{B}_{+}^{N\times K}$ denotes an $N\times K$ binary matrix, $k\in[1,K]$ denotes $k=1,\ldots,K$, and whenever complex matrix is concerned, transpose operation refers to conjugate transpose.

\section{The Ky Fan theorem} \label{kyFan}
The Ky Fan theorem \cite{Nakic} relates eigenvectors of a Hermitian matrix to the trace maximization problem of the matrix.
\begin{theorem} \label{theorem0}
The optimal value of the following problem:
\begin{equation}
\max_{\mathbf{X}^T\mathbf{X}=\mathbf{I}_K} \mathrm{tr}(\mathbf{X}^T\mathbf{H}\mathbf{X})
\label{eq4}
\end{equation}
is equal to $\sum_{k=1}^{K}\lambda_k$ if
\begin{equation}
\mathbf{X} = [\mathbf{u}_1,\ldots,\mathbf{u}_K]\mathbf{Q},
\label{eq5}
\end{equation}
where $\mathbf{H}\in\mathbb{C}^{N\times N}$ denotes a full rank Hermitian matrix with eigenvalues $\lambda_1\ge\ldots\ge\lambda_N\in\mathbb{R}_{+}$, $1\le K \le N$, $\mathbf{X}\in\mathbb{C}^{N\times K}$ denotes a unitary matrix, $\mathbf{I}_K$ denotes a $K\times K$ identity matrix, $\mathbf{u}_k\in\mathbb{C}^{N}$ denotes $k$-th eigenvector corresponds to $\lambda_k$, and $\mathbf{Q}\in\mathbb{C}^{K\times K}$ denotes an arbitrary unitary matrix.
\end{theorem}

The solution to eq.~\ref{eq4} is not unique since $\mathbf{X}$ remains equally good for arbitrary rotation and reflection due to the existence of unitary matrix $\mathbf{Q}$. However, since $[\mathbf{u}_1,\ldots,\mathbf{u}_K]$ is one of the optimal solution, setting $\mathbf{X}=[\mathbf{u}_1,\ldots,\mathbf{u}_K]$ eventually leads to the optimal value.

If $\mathbf{H}\leftarrow\mathbf{W}\in\mathbb{R}_{+}^{N\times N}$ where $\mathbf{W}$ denotes a symmetric af\mbox{}finity matrix induced from a graph, and $\mathbf{X}$ is constrained to be nonnegative while preserving the orthogonality, i.e., $\mathbf{X}^T\mathbf{X}=\mathbf{I}_K$, then problem in eq,~\ref{eq4} turns into $K$-way graph cuts problem. Therefore, the Ky Fan theorem can be viewed as a relaxed version of the graph cuts. This relationship explains the logic behind the spectral clustering, where an orthogonal nonnegative clustering indicator matrix is derived by computing the first $K$ eigenvectors of $\mathbf{W}$.

Eigenvectors of a matrix can be computed by using singular value decomposition (SVD). Hence, SVD will be discussed in the following section as many algorithms and software for computing SVD are available for use.

\section{Singular value decomposition} \label{svd}
SVD is a matrix decomposition technique that factorizes a matrix into a combination of left eigenvectors, right eigenvectors, and eigenvalues. SVD of a full rank matrix $\mathbf{A}\in\mathbb{C}^{M\times N}$ is defined as:
\begin{equation}
\mathbf{A} = \sum_{k=1}^{\min(M,N)} \sigma_k\mathbf{u}_k\mathbf{v}_k^T.
\label{eq1}
\end{equation}
Or, in a more compact form can be written as:
\begin{equation}
\mathbf{A} = \mathbf{U}\mathbf{\Sigma}\mathbf{V}^T,
\label{eq2}
\end{equation}
where $\mathbf{U}\in\mathbb{C}^{M\times M} = [\mathbf{u}_1,\ldots,\mathbf{u}_M]$ denotes an orthogonal matrix contains the left singular vectors of $\mathbf{A}$, $\mathbf{V}\in\mathbb{C}^{N\times N} = [\mathbf{v}_1,\ldots,\mathbf{v}_N]$ denotes an orthogonal matrix contains the right singular vectors of $\mathbf{A}$, and $\mathbf{\Sigma}\in\mathbb{R}_{+}^{M\times N}$ denotes a diagonal matrix contains the singular values $\sigma_1\ge\ldots\ge\sigma_{\min(M,N)}$ of $\mathbf{A}$ along its diagonal.

In practice, usually rank-$K$ approximation of $\mathbf{A}$ is used instead:
\begin{equation}
\mathbf{A}_K = \mathbf{U}_K\mathbf{\Sigma}_K \mathbf{V}_K^T,
\label{eq3}
\end{equation}
where usually $K\ll\min(M,N)$, $\mathbf{U}_K$ and $\mathbf{V}_K$ contain the first $K$ columns of $\mathbf{U}$ and $\mathbf{V}$ respectively, and $\mathbf{\Sigma}_K$ denotes a $K\times K$ principal submatrix of $\mathbf{\Sigma}$. According to Eckart-Young theorem, $\mathbf{A}_K$ is the closest rank-$K$ approximation of $\mathbf{A}$ \cite{Eckart}.

In the following section we show how to modify graph clustering objectives into trace maximization of corresponding symmetric matrices. And by relaxing the nonnegativity constraints, according to the Ky Fan theorem, clustering problems eventually become the tasks of finding the first $K$ eigenvectors of the matrices, which are exactly the SVD problems.

\section{Graph clustering} \label{dataClustering}
Graphs usually can be represented by symmetric, rectangular, or square af\mbox{}finity matrices. A collection of items connected by weighted edges describing similarities between item pairs like a friendship network can be modeled by a unipartite graph, then a symmetric af\mbox{}finity matrix can be induced from this graph. A collection of documents (and in general any bipartite dataset) can be modeled by a bipartite graph, and a term-by-document rectangular matrix containing (adjusted) frequencies of those terms in the documents can be constructed. And a square af\mbox{}finity matrix can be induced from a (unipartite) directed graph like WWW network.

Let $\mathcal{G}\left(\mathbf{A}\right)\equiv\mathcal{G}\left(\mathcal{V},\mathcal{E},\mathbf{A}\right)$ be the graph representation of a collection with $\mathcal{V}$ denotes the set of vertices, $\mathcal{E}$ denotes the set of edges connecting vertex pairs, and $\mathbf{A}$ denotes the induced af\mbox{}finity matrix. The $K$-way graph clustering is the problem of finding the best cuts on $\mathcal{G}\left(\mathbf{A}\right)$ that maximize within cluster association, or equivalently, minimize inter cluster cuts to produce $K$ clusters of $\mathcal{V}$. 

Here we state two assumptions to allow the graph cuts be employed in clustering.
\begin{assumption} \label{assumption1}
Let $e_{ij}$ be an edge connecting vertex $v_i$ to $v_j$, the weight value $|e_{ij}|$ denotes the similarity between $v_i$ and $v_j$ linearly, i.e., if $|e_{ij}| = n|e_{ik}|$ then $v_i$ is $n$ times more similar to $v_j$ than to $v_k$. And zero weight means no similarity.
\end{assumption}
Note that similarity term has many interpretations depending on the domain. For example, in the city road network the similarity can refer to the distance; the closer the distance between two points, the more similar those points are. And in the movie ratings, the similarity can refer to the number of common movies rated by the users.

\begin{assumption} \label{assumption2}
Graph clustering refers to hard clustering, i.e., for $\{\mathcal{V}_k\}_{k=1}^{K}\subset\mathcal{V}$, $\cup_{k=1}^{K}\mathcal{V}_k=\mathcal{V}$, and $\mathcal{V}_k\cap\mathcal{V}_l=\emptyset$ $\forall k\neq l$.
\end{assumption}
\begin{proposition} \label{propx}
Assumption \ref{assumption1} and \ref{assumption2} lead to the grouping of similar vertices in $\mathcal{G}\left(\mathbf{A}\right)$. 
\end{proposition}
\begin{proof}
Consider $\mathcal{G}\left(\mathbf{A}\right)$ to be clustered into $K$ groups by initial random assignments. Since assumption \ref{assumption1} guarantees $|e_{ij}|$ to be comparable, and assumption \ref{assumption2} guarantees each vertex to be assigned only to a single cluster, cluster assignment for $v_i$ ($z_{ik}$) can be found by finding a cluster's center that is most similar to $v_i$.
\begin{equation}
z_{ik} = \nut_{k}\,\max\left(\sum_{j_k=v_{j_k}\in\mathcal{V}_{k}}\frac{|e_{ij_k}|}{|\mathcal{V}_{k}|}\;\Bigg|\;k\in[1,K]\right),
\label{eqx}
\end{equation}
%\begin{equation}
%x_{ik} = \nut_{k}\,\max\left(\sum_{j_1=v_j\in\mathcal{V}_{1}}\frac{|e_{ij_1}|}{|\mathcal{V}_{1}|},\ldots,\sum_{j_K=v_j\in\mathcal{V}_{K}}\frac{|e_{ij_K}|}{|\mathcal{V}_{K}|}\right),
%\label{eqx}
%\end{equation}
where $|\mathcal{V}_{k}|$ denotes the size of cluster $k$. The objective in eq.~\ref{eqx} is the $K$-means clustering applied to $\mathcal{G}\left(\mathbf{A}\right)$, therefore leads to the grouping of similar vertices.
\end{proof}

Note that assumption \ref{assumption1} is an ideal situation which generally doesn't hold. For example, in bipartite representation of a term-by-document matrix, usually the relationships between term-document pairs are not linear to the corresponding term frequencies. Therefore, preprocessing steps (e.g., feature selection and term weighting) are usually necessary before applying the graph cuts. The preprocessing steps seem to be very crucial for obtaining good results \cite{Bach}, and many works are devoted to find more accurate similarity measures schemes \cite{Bach, Jin, Wu, Jain}. 

Even though the similarities have been reflected by the weights in (almost) linear fashion, a normalization scheme on $\mathbf{A}$ generally is preferable to produce balance-size clusters. In fact, normalized association/cuts objectives are proven to of\mbox{}fer better results compared to their unnormalized counterparts, ratio association/cuts objectives \cite{Shi, Yu, Dhillon}.

Table \ref{tablea} shows the most popular graph clustering objectives with the first two objectives are from the work of Dhillon et al.~\cite{Dhillon}. \emph{GWAssoc} (\emph{GWCuts}) refers to general weighted association (cuts), \emph{NAssoc} (\emph{NCuts}) refers to normalized association (cuts), and \emph{RAssoc} (\emph{RCuts}) refers to ratio association (cuts). Since all other objectives can be derived from \emph{GWAssoc} \cite{Dhillon}, we will only consider \emph{GWAssoc} for the rest of this paper.

\subsection{Unipartite graph clustering} \label{unipartite}
Unipartite graph is the framework for deriving a unified treatment for the three graphs, so we discuss it first. The following proposition summarizes the ef\mbox{}fort of Dhillon et al.~\cite{Dhillon} in providing a general unipartite graph clustering objective.

\begin{proposition} \label{proposition0}
Unipartite graph clustering can be stated in the trace maximization problem of a symmetric matrix.
\end{proposition}

\begin{proof}
Let $\mathbf{W}\in\mathbb{R}_{+}^{N\times N}$ be the symmetric af\mbox{}finity matrix induced from a unipartite graph, $K$-way partitioning on $\mathcal{G}\left(\mathbf{W}\right)$ using \textit{GWAssoc} can be found by:
\begin{equation}
\max\;J_u = \frac{1}{K}\sum_{k=1}^{K}\frac{z^T_k\mathbf{W}z_k}{z^T_k\mathbf{\Phi}z_k}
\label{eqa}
\end{equation}
where $\mathbf{\Phi}\in\mathbb{R}^{N\times N}_{+}$ denotes a diagonal matrix with $\Phi_{ii}$ associated with weight of $v_i$, and $\mathbf{z}_k\in\mathbb{B}^{N}_{+}$ denotes a binary indicator vector for cluster $k$ with its $i$-th entry is 1 if $v_i$ in cluster $k$, and 0 otherwise. 

The objective above can be rewritten more compactly in the trace maximization as:
\begin{align}
\max\;J_u &= \frac{1}{K}\text{tr}\left(\frac{\mathbf{Z}^T\mathbf{W}\mathbf{Z}}{\mathbf{Z}^T\mathbf{\Phi}\mathbf{Z}}\right) \nonumber \\
        &= \frac{1}{K}\text{tr}\left(\bar{\mathbf{Z}}^T\mathbf{\Phi}^{-1/2}\mathbf{W}\mathbf{\Phi}^{-1/2}\bar{\mathbf{Z}}\right)
\label{eqb}
\end{align}
where $\mathbf{Z}\in\mathbb{B}^{N\times K}_{+}=[\mathbf{z}_1,\ldots,\mathbf{z}_K]$ denotes the clustering indicator matrix, and $\bar{\mathbf{Z}}\in\mathbb{R}^{N\times K}_{+}=\mathbf{Z}/\left(\mathbf{Z}^T\mathbf{Z}\right)^{1/2}$ denotes its orthonormal version.
\end{proof}

By relaxing the strict nonnegativity constraints, i.e., allowing $\bar{\mathbf{Z}}$ to contain negative values while preserving its orthonormality, according to the Ky Fan theorem, the global optimum of $J_u$ can be obtained by assigning
\begin{equation}
\mathbf{\hat{Z}} = [\mathbf{u}_1,\ldots,\mathbf{u}_K]\mathbf{Q},
\label{eqc}
\end{equation}
where $\mathbf{u}_1,\ldots,\mathbf{u}_K\in\mathbb{C}^N$ denote the first $K$ eigenvectors of $\mathbf{\Phi}^{-1/2}\mathbf{W}\mathbf{\Phi}^{-1/2}$, $\mathbf{\hat{Z}}\in\mathbb{C}^{N\times K}$ denotes a relaxed version of $\mathbf{\bar{Z}}$, and $\mathbf{Q}\in\mathbb{R}^{K\times K}$ denotes an arbitrary orthonormal matrix. Hence, eq.~\ref{eqc} presents a tractable solution for NP-hard problem in eq.~\ref{eqb}.

The \emph{GWAsssoc} objective in eq.~\ref{eqb} can be replaced by any objective in table \ref{tablea} by substituting $\mathbf{W}$ and $\mathbf{\Phi}$ with corresponding af\mbox{}finity and weight matrices. Note that $\mathbf{I}$ denotes the identity matrix, $\mathbf{D}\in\mathbb{R}^{N\times N}_{+}$ denotes a diagonal matrix with its diagonal entries defined as $D_{ii}=\sum_j W_{ij}$, and $\mathbf{L}=\mathbf{D}-\mathbf{W}$ denotes the Laplacian of $\mathcal{G}\left(\mathbf{W}\right)$.

\begin{table}[t]
  \begin{center}
    \caption{\small Graph clustering objectives.}
    \centering
    \footnotesize{
    \begin{tabular}{|l|l|l|}
    \hline
   Objective & Af\mbox{}finity matrix & Weight matrix \\
    \hline
    \textit{GWAssoc} & $\mathbf{W}$ & $\mathbf{\Phi}$ \\
    \textit{GWCuts}  & $\mathbf{\Phi}-\mathbf{L}$ & $\mathbf{\Phi}$ \\
    \textit{NAssoc}  & $\mathbf{W}$ & $\mathbf{D}$ \\
    \textit{NCuts}   & $\mathbf{D}-\mathbf{L}$ & $\mathbf{D}$ \\
    \textit{RAssoc}  & $\mathbf{W}$ & $\mathbf{I}$ \\
    \textit{RCuts}   & $\mathbf{I}-\mathbf{L}$ & $\mathbf{I}$ \\
    \hline
    \end{tabular}}
    \label{tablea}
  \end{center}
\end{table}

\subsection{Bipartite graph clustering} \label{bipartite}
Bipartite graph clustering generally refers to the clustering of bipartite datasets---collections of items that are characterized by some shared features. A feature-by-item rectangular data matrix $\mathbf{A}\in\mathbb{R}^{M\times N}_{+}$ contains entries that describe the relationships between items and features.

Bipartite graph clustering can be done in two dif\mbox{}ferent ways; direct and indirect way. The former method applies the graph cuts directly to $\mathcal{G}\left(\mathbf{A}\right)$ resulting in partitions that contain both item and feature vertices. And the latter method first transforms $\mathcal{G}\left(\mathbf{A}\right)$ into an equivalent unipartite graph (either item or feature graph) by calculating similarities between vertex pairs from either item or feature set, and then applies the graph cuts on this unipartite graph. Both methods lead to symmetric af\mbox{}finity matrices, thus \emph{GWAssoc} objective can be applied as in the unipartite graph case equivalently.

\subsubsection{Direct treatment}
If \emph{GWAssoc} is applied to a bipartite graph, similar items will be grouped together with relevant features. This is known as simultaneous feature and item clustering or \emph{co-clustering}.

\theoremstyle{plain} 
\begin{proposition} \label{prop1}
Bipartite graph co-clustering can be stated in the trace maximization problem of a symmetric matrix.
\end{proposition}

\begin{proof}
Let $\mathbf{M}\in\mathbb{R}^{P\times P}_{+}$ ($P = M + N$) be the symmetric af\mbox{}finity matrix induced from a bipartite graph. $\mathbf{M}$ is defined as:
\begin{equation}
\mathbf{M} = \left[
 \begin{array}{cc}
  \mathbf{0}   & \mathbf{A} \\
  \mathbf{A}^T & \mathbf{0} \\
 \end{array} \right].
\label{eq6}
\end{equation}
Taking \textit{GWAssoc} as the objective, $K$-way co-clustering can be found by:
\begin{equation}
\max\;J_b = \frac{1}{K}\sum_{k=1}^{K}\frac{z^T_k\mathbf{M}z_k}{z^T_k\mathbf{\Phi}z_k}.
\label{eq7}
\end{equation}
Then, eq.~\ref{eq7} can be rewritten as:
\begin{equation}
\max\;J_b = \frac{1}{K}\text{tr}\left(\bar{\mathbf{Z}}^T\mathbf{\Phi}^{-1/2}\mathbf{M}\mathbf{\Phi}^{-1/2}\bar{\mathbf{Z}}\right),
\label{eqd}
\end{equation}
where $\mathbf{\Phi}\in\mathbb{R}^{P\times P}_{+}$, $\mathbf{z}_k\in\mathbb{B}^{P}_{+}$, and $\bar{\mathbf{Z}}\in\mathbb{R}^{P\times K}_{+}$ are defined equivalently as in the unipartite graph case.
\end{proof}

By relaxing the nonnegativity constraints on $\bar{\mathbf{Z}}$, the optimum value of eq.~\ref{eqd} can be found by computing the first $K$ eigenvectors of $\mathbf{\Phi}^{-1/2}\mathbf{M}\mathbf{\Phi}^{-1/2}$.

Instead of constructing $\mathbf{M}$ which is bigger and sparser than the original matrix $\mathbf{A}$, we provide a way to co-cluster bipartite graph directly from $\mathbf{A}$.

\theoremstyle{plain} 
\begin{theorem} \label{theorem2}
A relaxed solution to the bipartite graph co-clustering problem in eq.~\ref{eqd} can be found by computing the left and right eigenvectors of normalized version of $\mathbf{A}$.
\end{theorem}

\begin{proof}
Let
\begin{equation}
\mathbf{\bar{Z}}=\left[\begin{array}{c}\mathbf{\bar{X}}\\\mathbf{\bar{Y}}\end{array}\right]\,\mathrm{,}\;\mathrm{and}\;
\mathbf{\Phi}=\left[\begin{array}{cc}\mathbf{\Phi}_{1} & \mathbf{0}\\
\mathbf{0} & \mathbf{\Phi}_{2}\end{array}\right]
\label{eqe}
\end{equation}
be rearranged into two smaller matrices that correspond to $\mathbf{A}$ and $\mathbf{A}^T$ respectively. Then, eq.~\ref{eqd} can be rewritten as:
\begin{equation}
\max\;J_b = \frac{1}{K}\text{tr}\left(\left[\begin{array}{c}\mathbf{\bar{X}}\\\mathbf{\bar{Y}}\end{array}\right]^T\underbrace{\left[\begin{array}{cc}\mathbf{0} & \mathbf{\bar{A}}\\
\mathbf{\bar{A}}^T & \mathbf{0}\end{array}\right]}_{\mathbf{\bar{M}}}\left[\begin{array}{c}\mathbf{\bar{X}}\\\mathbf{\bar{Y}}\end{array}\right]\right),
\label{eqf}
\end{equation}
where $\mathbf{\bar{A}}=\mathbf{\Phi}_{1}^{-1/2}\mathbf{A}\mathbf{\Phi}_{2}^{-1/2}$. Denoting $\mathbf{\hat{X}}\in\mathbb{C}^{M\times K}$ and $\mathbf{\hat{Y}}\in\mathbb{C}^{N\times K}$ as the relaxed version of $\mathbf{\bar{X}}$ and $\mathbf{\bar{Y}}$, by the Ky Fan theorem, the global optimum solution to eq.~\ref{eqf} is given by the first $K$ eigenvectors of $\mathbf{\bar{M}}$:
\begin{equation}
\left[\begin{array}{c}\mathbf{\hat{X}}\\\mathbf{\hat{Y}}\end{array}\right]=
\left[\begin{array}{c}\mathbf{\hat{x}}_1,\ldots,\mathbf{\hat{x}}_K\\
\mathbf{\hat{y}}_1,\ldots,\mathbf{\hat{y}}_K\end{array}\right]\mathbf{Q}.
\label{eqg}
\end{equation}
Therefore, 
\begin{equation}
\left[\begin{array}{cc}\mathbf{0} & \mathbf{\bar{A}}\\
\mathbf{\bar{A}}^T & \mathbf{0}\end{array}\right]\left[\begin{array}{c}\mathbf{\hat{x}}_k\\
\mathbf{\hat{y}}_k\end{array}\right] =
\lambda_k\left[\begin{array}{c}\mathbf{\hat{x}}_k\\
\mathbf{\hat{y}}_k\end{array}\right],
\label{eqh}
\end{equation}
where $k\in[1,K]$ and $\lambda_k$ denotes $k$-th eigenvalue of $\mathbf{\bar{M}}$. Then,
\begin{align}
\mathbf{\bar{A}}\mathbf{\hat{y}}_k &= \lambda_k\mathbf{\hat{x}}_k,\,\; \text{and} \label{eqyxk}\\
\mathbf{\bar{A}}^T\mathbf{\hat{x}}_k &= \lambda_k\mathbf{\hat{y}}_k. \label{eqxyk}
\end{align}
Thus, a relaxed global optimum solution to the problem in eq.~\ref{eqd} can be found by computing the first $K$ left and right eigenvectors of $\mathbf{\bar{A}}$.
\end{proof}

Theorem \ref{theorem2} generalizes the work of Dhillon \cite{Dhillon0} where the author only gives a theoretical explanation for 2-way bipartite graph co-clustering. And the multipartitioning algorithm proposed by the author \cite{Dhillon0} that derived from the bipartitioning algorithm by induction, now has a theoretical explanation.

The following theorem gives a support for an interesting claim in co-clustering: row clustering implies column clustering and vice versa.

\theoremstyle{plain} 
\begin{theorem} \label{theorem3}
Solving simultaneous row and column clustering is equivalent to solving row and column clustering separately, and consequently, row clustering implies column clustering and vice versa.
\end{theorem}

\begin{proof}
By substituting $\mathbf{\hat{y}}_k$ from eq.~\ref{eqxyk} into eq.~\ref{eqyxk}, and similarly, substituting $\mathbf{\hat{x}}_k$ from eq.~\ref{eqyxk} into eq.~\ref{eqxyk}, we get:
\begin{align}
\mathbf{\bar{A}}\mathbf{\bar{A}}^T\mathbf{\hat{x}}_k &= \lambda_k^2\mathbf{\hat{x}}_k,\,\; \text{and} \label{xxk} \\
\mathbf{\bar{A}}^T\mathbf{\bar{A}}\mathbf{\hat{y}}_k &= \lambda_k^2\mathbf{\hat{y}}_k \label{yyk},
\end{align}
where $\mathbf{\bar{A}}\mathbf{\bar{A}}^T$ and $\mathbf{\bar{A}}^T\mathbf{\bar{A}}$ respectively denote row and column af\mbox{}finity matrices. After some manipulations, we get:
\begin{align}
\max\;\mathrm{tr}\left(\mathbf{\hat{X}}^T\mathbf{\bar{A}}\mathbf{\bar{A}}^T\mathbf{\hat{X}}\right) &=\sum_{k=1}^{K}\lambda_k^2,\,\; \text{and} \label{xxkk} \\
\max\;\mathrm{tr}\left(\mathbf{\hat{Y}}^T\mathbf{\bar{A}}^T\mathbf{\bar{A}}\mathbf{\hat{Y}}\right) &=\sum_{k=1}^{K}\lambda_k^2 \label{yykk},
\end{align}
where $\mathbf{\hat{X}}=[\mathbf{\hat{x}}_1,\ldots,\mathbf{\hat{x}}_K]$ and $\mathbf{\hat{Y}}=[\mathbf{\hat{y}}_1,\ldots,\mathbf{\hat{x}}_K]$ respectively denote the relaxed row and column clustering indicator matrices. As shown above, $\mathbf{\hat{X}}$ and $\mathbf{\hat{Y}}$ can be computed separately, and since $\mathbf{\hat{Y}}$ can be derived from $\mathbf{\hat{X}}$ and vice versa (see eq.~\ref{eqyxk} and eq.~\ref{eqxyk}), row clustering implies column clustering and vice versa.
\end{proof}

Theorem \ref{theorem2} provides a "shortcut" to computing $\mathbf{\hat{X}}$ and $\mathbf{\hat{Y}}$ which are usually be constructed by computing the first $K$ eigenvectors of $\mathbf{\bar{A}}\mathbf{\bar{A}}^T$ and $\mathbf{\bar{A}}^T\mathbf{\bar{A}}$ respectively.
\begin{theorem} \label{theorem3a}
$\mathbf{\hat{X}}$ and $\mathbf{\hat{Y}}$ can be constructed by computing the first $K$ left and right eigenvectors of $\mathbf{\bar{A}}$.
\end{theorem}
\begin{proof}
As shown in the proof of theorem \ref{theorem3}, $\mathbf{\hat{X}}=[\mathbf{\hat{x}}_1,\ldots,\mathbf{\hat{x}}_K]$ and $\mathbf{\hat{Y}}=[\mathbf{\hat{y}}_1,\ldots,\mathbf{\hat{y}}_K]$, where according to the proof of theorem \ref{theorem2}, $\mathbf{\hat{x}}_k$ and $\mathbf{\hat{y}}_k$ are the $k$-th left and right eigenvectors of $\mathbf{\bar{A}}$.
\end{proof}

\subsubsection{Indirect treatment}
There are cases where the data points are inseparable in the original space or clustering can be done more effectively by first transforming $\mathbf{A}$ into a corresponding symmetric matrix $\mathbf{V}\in\mathbb{R}_{+}^{N\times N}$ (we assume item clustering for the rest of this subsection, feature clustering can be done similarly). Then the graph cuts can be applied to $\mathcal{G}(\mathbf{V})$ to obtain the item clustering.

There are two common approaches to learn $\mathbf{V}$ from $\mathbf{A}$. The first approach is to use kernel functions. Table \ref{table2} lists the most widely used kernel functions according to Dhillon et al.~\cite{Dhillon} with $\mathbf{a}_i$ is $i$-th column of $\mathbf{A}$, and the unknown parameters ($c,d,\alpha$, and $\theta$) are either directly determined based on previous experiences or learned from sample datasets. 

The second approach is to make no assumption about the data domain nor the possible similarity structure between item pairs. $\mathbf{V}$ is learned directly from the data, thus avoiding some inherent problems associated with the first approach, e.g., (1) no standard in choosing the kernel function and (2) similarities between item pairs are computed independently without considering interactions among items. Some recent works on this approach can be found in \cite{Jin, Wu, Jain}.

\begin{table}[t]
 \renewcommand{\arraystretch}{1.3}
  \begin{center}
    \caption{\small Examples of popular kernel functions \cite{Dhillon}.}
    \centering
    \footnotesize{
    \begin{tabular}{|l|l|}
    \hline
    Polynomial kernel & $\kappa(\mathbf{a}_i,\mathbf{a}_j)=(\mathbf{a}_i\cdot\mathbf{a}_j+c)^d$ \\
    Gaussian kernel   & $\kappa(\mathbf{a}_i,\mathbf{a}_j)=\mathrm{exp}(-\|\mathbf{a}_i-\mathbf{a}_j\|^2/2\alpha^2)$ \\
    Sigmoid kernel    & $\kappa(\mathbf{a}_i,\mathbf{a}_j)=\mathrm{tanh}(c(\mathbf{a}_i\cdot\mathbf{a}_j)+\theta)$ \\
    \hline
    \end{tabular}}
    \label{table2}
  \end{center}
\end{table}

\begin{proposition} \label{proposition3}
Clustering on $\mathcal{G}(\mathbf{V})$ can be stated in the trace maximization of $\mathbf{V}$.
\end{proposition}

\begin{proof}
If the first approach to be used, entries of $\mathbf{V}$ can be determined using a kernel function,
\begin{equation}
V_{ij} = \left\{
  \begin{array}{rl}
    \kappa(\mathbf{a}_i,\mathbf{a}_j) & \text{if } i\neq j \\
    0 & \text{if } i=j
  \end{array} \right.
\label{eqwa}
\end{equation}
Similarly, if the second approach to be used, $\mathbf{V}$ can be learned directly from the data. Then, by using \emph{GWAssoc} as the objective, $K$-way clustering on $\mathcal{G}\left(\mathbf{V}\right)$ can be computed by:
\begin{equation}
\max\;J_b = \frac{1}{K}\text{tr}\left(\bar{\mathbf{Z}}^T\mathbf{\Phi}^{-1/2}\mathbf{V}\mathbf{\Phi}^{-1/2}\bar{\mathbf{Z}}\right)
\label{eqbw}
\end{equation}
where $\mathbf{\bar{Z}}$ and $\mathbf{\Phi}$ are defined equivalently as in the unipartite graph case.
\end{proof}
If asymmetric metrics like Bregman divergences are used as the kernel functions, the resulting $\mathbf{V}'$ will be asymmetric. Accordingly, $\mathcal{G}(\mathbf{V}')$ is a directed graph, and therefore it must be treated as a directed graph.

%It is stated in \cite{Xu} that document clustering using latent semantic indexing (LSI)---a method that uses the left and right eigenvectors of the term-by-document matrix $\mathbf{A}$ to determine the clustering---often produces poor results compared to the clustering by using graph clustering objectives on $\mathcal{G}(\mathbf{V})$. However, as shown in theorem \ref{theorem2} and \ref{theorem3}, LSI should produce a relaxed global optimum solution. Therefore the poor results stated in \cite{Xu} are not due to the method itself, but perhaps because (1) the entries of $\mathbf{A}$ do not linearly describe the relationships between corresponding term-document pairs, (2) LSI is applied to the unnormalized $\mathbf{A}$, or (3) the postprocessing step (usually by using $K$-means algorithm) to induce $\mathbf{\bar{X}}$ and $\mathbf{\bar{Y}}$ from $\mathbf{\hat{X}}$ and $\mathbf{\hat{Y}}$ leads to the poor result.

\subsection{Directed graph clustering} \label{directed}
The researches on directed graph clustering come from complex network studies conducted mainly by physicists. Dif\mbox{}ferent from conventional method of ignoring the edge directions, complex network researchers preserve this information in their proposed methods. As shown in \cite{Leicht, Kim}, accomodating it can be very useful in improving clustering quality. In some cases, ignoring the edge directions can lead to the clusters detection failure \cite{Kim1}.

The directed graph clustering usually is done by mapping the original square af\mbox{}finity matrix into another square matrix which entries are adjusted to emphasize the importance of the edge directions. Some mapping functions can be found in, e.g., \cite{Leicht, Kim, Kim1}. To make use of the available clustering methods for unipartite graph, some works \cite{Leicht, Kim1} construct a symmetric matrix representation of the directed graph without ignoring the edge directions.

Here we describe the directed graph clustering by naturally following the previous discussions on the unipartite and bipartite graph cases.

\begin{proposition} \label{proposition4}
Directed graph clustering can be stated in the trace maximization problem of a symmetric matrix.
\end{proposition}

\begin{proof}
Let $\mathbf{B}\in\mathbb{R}_{+}^{N\times N}$ be the af\mbox{}finity matrix induced from a directed graph, and $\mathbf{\Phi}_i$ and $\mathbf{\Phi}_o$ be diagonal weight matrices associated with indegree and outdegree of vertices in $\mathcal{G}\left(\mathbf{B}\right)$ respectively. We define a diagonal weight matrix of $\mathcal{G}\left(\mathbf{B}\right)$ with:
\begin{equation}
\mathbf{\Phi}_{io} = \sqrt{\mathbf{\Phi}_{i}\mathbf{\Phi}_{o}}.
\end{equation}
Since both rows and columns of $\mathbf{B}$ correspond to the same set of vertices with the same order, the row and column clustering indicator matrices are the same, matrix $\mathbf{\bar{Z}}$. By using \emph{GWAssoc}, $K$-way clustering on $\mathcal{G}\left(\mathbf{B}\right)$ and $\mathcal{G}\left(\mathbf{B}^T\right)$ can be found by:
\begin{align}
\max\;J_{d1} &= \frac{1}{K}\text{tr}\left(\bar{\mathbf{Z}}^T\mathbf{\Phi}_{io}^{-1/2}\mathbf{B}\mathbf{\Phi}_{io}^{-1/2}\bar{\mathbf{Z}}\right),\;\text{and} \label{eqdb} \\
\max\;J_{d2} &= \frac{1}{K}\text{tr}\left(\bar{\mathbf{Z}}^T\mathbf{\Phi}_{io}^{-1/2}\mathbf{B}^T\mathbf{\Phi}_{io}^{-1/2}\bar{\mathbf{Z}}\right)
\label{eqdbt}
\end{align}
respectively. By adding the two objectives above, we obtain:
\begin{equation}
\max\;J_d = \frac{1}{K}\text{tr}\left(\bar{\mathbf{Z}}^T\mathbf{\Phi}_{io}^{-1/2}\left(\mathbf{B}+\mathbf{B}^T\right)\mathbf{\Phi}_{io}^{-1/2}\bar{\mathbf{Z}}\right),
\label{eqdc}
\end{equation}
which is the trace maximization problem of a symmetric matrix $\mathbf{\Phi}_{io}^{-1/2}\left(\mathbf{B}+\mathbf{B}^T\right)\mathbf{\Phi}_{io}^{-1/2}$.
\end{proof}
The directed graph clustering raises an interesting issue in the weight matrix formulation which doesn't appear in the unipartite and bipartite graph cases as the edges are undirected. As explained in the original work \cite{Dhillon}, $\mathbf{\Phi}$ is introduced with two purposes: \emph{first} to provide a general form of graph cuts objective which other objectives can be derived from it, and \emph{second} to provide compatibility with weighted kernel $K$-means objective so that eigenvector-free $K$-means algorithm can be utilized to solve the graph cuts problem.

However, as information of the edge directions appears, defining a weight for each vertex is no longer adequate. To see the reason, let's apply \emph{NAssoc} to $\mathcal{G}\left(\mathbf{B}\right)$ and $\mathcal{G}\left(\mathbf{B}^T\right)$. By using table \ref{tablea}:
\begin{align}
\max\;J_{d1} &= \frac{1}{K}\text{tr}\left(\bar{\mathbf{Z}}^T\mathbf{D}^{-1/2}\mathbf{B}\mathbf{D}^{-1/2}\bar{\mathbf{Z}}\right)\;\,\text{and} \label{eqdbx} \\
\max\;J_{d2} &= \frac{1}{K}\text{tr}\left(\bar{\mathbf{Z}}^T\mathbf{D^{\ast}}^{-1/2}\mathbf{B}^T\mathbf{D^{\ast}}^{-1/2}\bar{\mathbf{Z}}\right),
\label{eqdbtx}
\end{align}
where $D$ and $D^{\ast}$ are diagonal weight matrices with $D_{ii}=\sum_jB_{ij}$ and $D_{ii}^{\ast}=\sum_iB_{ij}$ respectively. But now $J_{d1}+J_{d2}$ won't end up in a nice trace maximization of a symmetric matrix as in eq.~\ref{eqdc}. Therefore, we cannot apply the Ky Fan theorem to find a relaxed global optimum solution.

This motivates us to define a more general form of the weight matrix, $\mathbf{\Phi}_{io}$, which allows directed graph clustering be stated in the trace maximization of a symmetric matrix, yet still turns into $\mathbf{\Phi}$ if the corresponding af\mbox{}finity matrix is symmetric. 

In the case of \emph{NAssoc} and \emph{NCuts}, $\mathbf{\Phi}_{i}$ and $\mathbf{\Phi}_{o}$ are defined as:
\begin{align}
\mathbf{\Phi}_i&=\text{diag}\left(\sum_{i}B_{i1},\ldots,\sum_{i}B_{iN}\right) \label{eqphii}\;\,\text{and} \\
\mathbf{\Phi}_o&=\text{diag}\left(\sum_{j}B_{1j},\ldots,\sum_{j}B_{Nj}\right). \label{eqphio}
\end{align}
Note that there is no need to define weight matrix for \emph{RAssoc} and \emph{RCuts} since $\mathbf{I}$ is used.

\section{Extension to the Ky Fan Theorem}
Theorem \ref{theorem2} implies an extension to the Ky Fan theorem for more general rectangular complex matrix.

\begin{theorem} \label{theorem4}
The optimal value of the following problem:
\begin{equation}
\max_{\mathbf{X}^T\mathbf{X}=\mathbf{Y}^T\mathbf{Y}=\mathbf{I}_K} \mathrm{tr}(\mathbf{X}^T\mathbf{R}\mathbf{Y}),
\label{eqkf}
\end{equation}
is equal to $\sum_{k=1}^{K}\lambda_k$ if 
\begin{align}
\mathbf{X} &= [\mathbf{x}_1,\ldots,\mathbf{x}_K]\mathbf{Q},\;\,\text{and}\label{eqxxx}\\
\mathbf{Y} &= [\mathbf{y}_1,\ldots,\mathbf{y}_K]\mathbf{Q}\label{eqyyy}
\end{align}
where $\mathbf{R}\in\mathbb{C}^{M\times N}$ denotes a full rank rectangular complex matrix with eigenvalues $\lambda_1\ge\ldots\ge\lambda_{\min(M,N)}\in\mathbb{R}_{+}$, $0\le K\le\min(M,N)$, $\mathbf{X}\in\mathbb{C}^{M\times K}$ and $\mathbf{Y}\in\mathbb{C}^{N\times K}$ denote unitary matrices, $\mathbf{x}_k$ and $\mathbf{y}_k$ ($k\in[1,K]$) respectively denote $k$-th left and right eigenvectors  correspond to $\lambda_k$, and $\mathbf{Q}\in\mathbb{C}^{K\times K}$ denotes an arbitrary unitary matrix.
\end{theorem}

\begin{proof}
Eq.~\ref{eqkf} can be rewritten as:
\begin{equation}
\max_{\mathbf{X}^T\mathbf{X}=\mathbf{Y}^T\mathbf{Y}=\mathbf{I}_K}\frac{1}{2} \mathrm{tr}\left(\left[\begin{array}{c}\mathbf{X}\\\mathbf{Y}\end{array}\right]^T\underbrace{\left[\begin{array}{cc}\mathbf{0} & \mathbf{R}\\
\mathbf{R}^T & \mathbf{0}\end{array}\right]}_{\mathbf{\Psi}}\left[\begin{array}{c}\mathbf{X}\\\mathbf{Y}\end{array}\right]\right).
\label{eqkkkk}
\end{equation}
Since $\mathbf{\Psi}$ is a Hermitian matrix, by the Ky Fan theorem, the global optimum solution is given by the first $K$ eigenvectors of $\mathbf{\Psi}$:
\begin{equation}
\left[\begin{array}{c}\mathbf{X}\\\mathbf{Y}\end{array}\right]=
\left[\begin{array}{c}\mathbf{x}_1,\ldots,\mathbf{x}_K\\
\mathbf{y}_1,\ldots,\mathbf{y}_K\end{array}\right]\mathbf{Q}.
\label{eqfff}
\end{equation}
By following the proof of theorem \ref{theorem2}, it can be shown that $\mathbf{x}_1,\ldots,\mathbf{x}_K$ and $\mathbf{y}_1,\ldots,\mathbf{y}_K$ are the first $K$ left and right eigenvectors of $\mathbf{R}$.
\end{proof}

Interestingly, theorem \ref{theorem4} can also be proven by using the SVD definition.
\begin{proof}
Without loosing generality, let assume $N\leq M$
\begin{align}
\mathbf{R}=&\mathbf{U}\mathbf{\Sigma}\mathbf{V}^T \nonumber \\
          =&\mathbf{U}_{1,\ldots,K}\mathbf{\Sigma}_{1,\ldots,K}\mathbf{V}_{1,\ldots,K}^T + \nonumber \\
           &\mathbf{U}_{K+1,\ldots,N}\mathbf{\Sigma}_{K+1,\ldots,N}\mathbf{V}_{K+1,\ldots,N}^T,
\end{align}
where $\mathbf{U}$ and $\mathbf{V}$ defined as in section \ref{svd}, $\mathbf{U}_{a,\ldots,b}$ and  $\mathbf{V}_{a,\ldots,b}$ denote matrices built by taking column $a$ to $b$ from $\mathbf{U}$ and $\mathbf{V}$ respectively, and $\mathbf{\Sigma}_{a,\ldots,b}=\text{diag}\left[\lambda_a,\ldots,\lambda_b\right]$. Then,
\begin{equation}
\mathbf{U}_{1,\ldots,K}^T\mathbf{R}\mathbf{V}_{1,\ldots,K} = \mathbf{\Sigma}_{1,\ldots,K},
\end{equation}
or more conveniently,
\begin{equation}
\mathbf{U}_K^T\mathbf{R}\mathbf{V}_K = \mathbf{\Sigma}_K.
\end{equation}
Therefore,
\begin{equation}
\mathrm{tr}\left(\mathbf{U}_K^T\mathbf{R}\mathbf{V}_K\right) = \sum_{k=1}^K \lambda_k = \max_{\mathbf{X}^T\mathbf{X}=\mathbf{Y}^T\mathbf{Y}=\mathbf{I}_K} \mathrm{tr}(\mathbf{X}^T\mathbf{R}\mathbf{Y}).
\label{eqttt}
\end{equation}
\end{proof}

Theorem \ref{theorem4} is the general form of theorem \ref{theorem2} and gives a theoretical support for directly applying the graph cuts on the data matrix $\mathbf{A}\in\mathbb{R}_{+}^{M\times N}$ to get simultaneous row and column clustering:
\begin{equation}
\max_{\mathbf{\bar{X}}^T\mathbf{\bar{X}}=\mathbf{\bar{Y}}^T\mathbf{\bar{Y}}=\mathbf{I}_K} \mathrm{tr}(\mathbf{\bar{X}}^T\mathbf{A}\mathbf{\bar{Y}}),
\label{eqxay}
\end{equation}
where $\mathbf{\bar{X}}\in\mathbb{R}_{+}^{M\times K}$ and $\mathbf{\bar{Y}}\in\mathbb{R}_{+}^{N\times K}$ denote the row and column clustering indicator matrices respectively.

\section{Related works} \label{relatedWorks}
Zha et al.~\cite{Zha} and Ding et al.~\cite{Ding} mention the Ky Fan theorem in their discussions on the spectral clustering. However, the role of the theorem in the spectral clustering can be easily overlooked as it is not clearly described.

The equivalences between $K$-means clustering and several graph cuts objectives to the trace maximization objectives are well-known facts in the spectral clustering researches as many papers discuss about it with exception for the directed graph case, as this problem arises from complex network researches. Some representative works are \cite{Zha, Ding, Yu, Dhillon, Dhillon2}.

Leicht et al.~\cite{Leicht} discuss how to extend the so-called modularity---which is equivalent to the graph cuts objective---of unipartite graph to directed graph. They form an asymmetric modularity matrix $\mathbf{B^{\ast}}\in\mathbb{R}_{+}^{N\times N}$ by applying modularity function to emphasizes the importance of the edge directions to the original asymmetric af\mbox{}finity matrix $\mathbf{B}\in\mathbb{R}_{+}^{N\times N}$, and then transform $\mathbf{B^{\ast}}$ into a symmetric matrix by adding $\mathbf{B^{\ast}}$ to its transpose. The clustering is done by calculating the first $K$ eigenvectors of this symmetric matrix. This is equivalent to applying \emph{RAssoc} to $(\mathbf{B^{\ast}}+\mathbf{B^{\ast}}^T)$.

%Kim et al.~\cite{Kim} propose a new modularity function for mapping the square af\mbox{}finity matrix induced from the directed graph into a square modularity matrix inspired by Google's PageRank algorithm. The clustering itself is done by using simulated annealing algorithm on the modularity matrix.

Kim et al.~\cite{Kim1} propose a method for transforming the af\mbox{}finity matrix induced from a directed graph into a symmetric matrix without ignoring the edge directions. So, clustering algorithms built for unipartite graph can be applied unchanged.

\section{A note on spectral clustering algorithms} \label{note}
There are many spectral clustering algorithms available. They are dif\mbox{}ferent in many aspects, from the chosen af\mbox{}finity matrices to the postprocessing methods to derive clustering from eigenvectors. According to Luxburg \cite{Luxburg}, the most popular ones are algorithms by Shi et al.~\cite{Shi} and by Ng et al.~\cite{Ng}, with the former is more favorable because the computed eigenvectors are more related to the clustering indicator vectors. 

Here we like to note that according to Dhillon et al.~\cite{Dhillon}, a state-of-the-art spectral clustering algorithm based on the work of Yu et al.~\cite{Yu} empirically performed the best among various spectral algorithms that were tested in the terms of optimizing the objective function values. Furthermore, the multilevel algorithm proposed in \cite{Dhillon}---which exploits the equivalences of various graph clustering objectives to weighted kernel $K$-means objective to eliminate the need for eigenvectors computation---shows very promising results which while moderately improving clustering quality, drastically improving computational speed (up to 2000 times faster than the spectral method) and memory usage.

\section{Conclusion} \label{conclusion}

We presented a concise explanation on the logic behind the spectral clustering. Unlike $K$-means clustering and graph cuts which are very intuitive and straightforward, the spectral clustering tends to be incomprehensible. By using the Ky Fan theorem, we showed that the spectral clustering has a simple explanation and is also intuitive. 

We showed how to treat $K$-way clustering on unipartite, bipartite and directed graphs as the trace maximization problems on the corresponding symmetric matrices, thus a unified treatment can be applied to those graphs.

In bipartite graph, we proved that the co-clustering can be obtained by computing the left and right eigenvectors of the corresponding feature-by-item data matrix, thus generalizing the result of Dhillon \cite{Dhillon0} and providing a theoretical basis for spectral co-clustering algorithms proposed in, e.g., \cite{Dhillon0, Klugar}. We also proved that solving simultaneous row and column clustering is equivalent to solving row and column clustering separately, thus giving a theoretical support for the claim: ``column clustering implies row clustering and vice versa'', and then gave a "shortcut" to compute the row and column clustering indicator matrices.

In directed graph, we described a new clustering objective by following the discussions on unipartite and bipartite graphs naturally.

By extending theorem \ref{theorem2} to complex domain, we generalized the Ky Fan theorem to rectangular complex matrix. The second proof of theorem \ref{theorem4} shows that this theorem is a corollary of the SVD formulation, and thus the Ky Fan theorem and its general form are the corollaries of the Eckart-Young theorem.

We must note that, however, as the mathematics behind the spectral clustering has a long story (the Ky Fan theorem itself was proposed in 50's), it is probable that the contributions in this paper are not new, or can be derived easily from other well-established facts, theorems, or definitions.

\end{document}